\documentclass[conference]{IEEEtran}
\IEEEoverridecommandlockouts
\usepackage{cite}
\usepackage{comment}
\usepackage{amsmath,amssymb,amsfonts}
\usepackage{algorithm}
\usepackage{algorithmic}
\newtheorem{definition}{Definition}
\usepackage{graphicx}
\usepackage{textcomp}
\usepackage{xcolor}
\usepackage{float}
\usepackage{multirow}
\usepackage{adjustbox}
\usepackage{cleveref}
\usepackage{subcaption}
\usepackage{flushend}
\usepackage[caption=false]{subfig}
\def\BibTeX{{\rm B\kern-.05em{\sc i\kern-.025em b}\kern-.08em
    T\kern-.1667em\lower.7ex\hbox{E}\kern-.125emX}}
\begin{document}

\title{Privacy Drift: Evolving Privacy Concerns in Incremental Learning}
\author{
    \IEEEauthorblockN{
    Sayyed Farid Ahamed\IEEEauthorrefmark{1},
    Soumya Banerjee\IEEEauthorrefmark{1}\IEEEauthorrefmark{3},
    Sandip Roy\IEEEauthorrefmark{1}\IEEEauthorrefmark{3},
    Aayush Kapoor\IEEEauthorrefmark{1},
    Marc Vucovich\IEEEauthorrefmark{2},
    }
    \IEEEauthorblockN{ 
    Kevin Choi\IEEEauthorrefmark{2}, 
    Abdul Rahman\IEEEauthorrefmark{2},
    Edward Bowen\IEEEauthorrefmark{2},
    Sachin Shetty\IEEEauthorrefmark{1}
    }
    \IEEEauthorblockA{
    \IEEEauthorrefmark{1}Center for Secure \& Intelligent Critical Systems, Old Dominion University, Virginia, USA \\
    \IEEEauthorrefmark{3}School of Cybersecurity, Old Dominion University, Virginia, USA
    \\\{saham001, s1banerj, sroy, akapo004, sshetty\}@odu.edu}
    \IEEEauthorblockA{
    \IEEEauthorrefmark{2}Deloitte \& Touche LLP
    \\\{mvucovich, kevchoi, abdulrahman, edbowen\}@deloitte.com}
}

\maketitle

\begin{abstract}
In the evolving landscape of machine learning (ML), Federated Learning (FL) presents a paradigm shift towards decentralized model training while preserving user data privacy. This paper introduces the concept of ``privacy drift", an innovative framework that parallels the well-known phenomenon of concept drift. While concept drift addresses the variability in model accuracy over time due to changes in the data, privacy drift encapsulates the variation in the leakage of private information as models undergo incremental training. By defining and examining privacy drift, this study aims to unveil the nuanced relationship between the evolution of model performance and the integrity of data privacy. Through rigorous experimentation, we investigate the dynamics of privacy drift in FL systems, focusing on how model updates and data distribution shifts influence the susceptibility of models to privacy attacks, such as membership inference attacks (MIA). Our results highlight a complex interplay between model accuracy and privacy safeguards, revealing that enhancements in model performance can lead to increased privacy risks. We provide empirical evidence from experiments on customized datasets derived from CIFAR-100 (Canadian Institute for Advanced Research, 100 classes), showcasing the impact of data and concept drift on privacy. This work lays the groundwork for future research on privacy-aware machine learning, aiming to achieve a delicate balance between model accuracy and data privacy in decentralized environments.
\end{abstract}

\begin{IEEEkeywords}
Federated Learning (FL),  Membership Inference Attack (MIA), Privacy Drift, Data Drift, Concept Drift, Model Drift, Incremental Learning.
\end{IEEEkeywords}

\section{Introduction}

In the realm of machine learning, particularly within the context of Federated Learning (FL), the balance between prediction accuracy and data privacy emerges as a critical frontier of research. This paper introduces the concept of  ``privacy drift'', a phenomenon analogous to the well-known phenomenon of concept drift.

Privacy drift is defined as the gradual increase in a model's vulnerability to information leakage over its training lifecycle. This phenomenon arises from continuous data updates or model modifications, which can unintentionally alter the ease with which an adversary can extract private information from the model.
Concept drift describes the variability in model accuracy as new data is introduced over time. Similarly, privacy drift encapsulates the variation in the leakage of private information under the same conditions. This exploration is rooted in the premise that as ML models evolve with incremental training \cite{wu2019large}, \cite{schlimmer1986incremental}, the interplay between accuracy and privacy is dynamic. By defining and examining privacy drift, this paper unveils the nuanced relationship between the evolution of model performance and the integrity of data privacy, offering novel insights into the challenges and strategies for managing privacy within incrementally updating learning systems.





In this paper, we aim to investigate and explore the inherent privacy drift in the context of incremental learning. The main contributions of this paper are as follows:\\
\noindent
\textbf{1. Introduction and Formalization of Privacy Drift}:
We define and formalize ``privacy drift" in FL, establishing a new framework to understand the variation in private information leakage as models evolve with incremental training.\\
\noindent
\textbf{2. Empirical Evidence of Privacy Drift}:
Through experiments on customized CIFAR-100 dataset, we provide empirical evidence showing how data distribution shifts and model updates affect model susceptibility to privacy attacks, such as membership inference attacks (MIA) \cite{banerjee2024mia}, \cite{shokri2017membership}.


The rest of the paper is organized as follows. In \Cref{background}, we introduce the relevant theoretical background and present a brief literature survey. \Cref{threat} defines the threat model, describing the goals, knowledge, and capabilities of the attacker and the defender. \Cref{analysis} provides an analysis of the concept of privacy drift. Experimental results and discussions are provided in \Cref{results}. Finally, we conclude our work and discuss future research scope in \Cref{conclusion}.

\section{Background}\label{background}

The security, performance, and effectiveness of data-centric operations are strongly dependent on the accuracy of a ML model \cite{flynn2019deepview}, \cite{arzani2020privateeye}. Declining accuracy may have serious repercussions, including noticeable system inefficiencies and performance loss. Drops in performance are often caused by \textbf{concept drift} or \textbf{data drift} \cite{ackerman2021automatically}, \cite{lu2018learning}.

\subsection{Data Drift}

In contrast to machine learning (ML) applications like image classification \cite{russakovsky2015imagenet} and natural language processing (NLP) \cite{chowdhary2020natural} (where the data often originates from a stable sample space), the majority of system applications involve data that is intrinsically temporal and subject to change over time \cite{ackerman2021automatically}. An inconsistency between training data from the \textit{past} and test data from the \textit{future} causes data drift issues (also termed as `covariate drift' or `feature drift'). Data drift occurs when the distribution of the input data changes between the training and testing phases.


The ML community identifies two main types of data drift problems \cite{moreno2012unifying}. \textit{First}, covariate drift or feature drift, that stems from a change in the distribution of independent variables. For example, a shift in the age distribution can impact the accuracy of a model \cite{xu2021concept}. \textit{Second}, prior probability shift or label drift, occurs when the distribution of dependent variables changes. An example is the varying proportions of fraudulent transactions in a fraud detection model, which can lead to performance degradation \cite{bai2022adapting}.

\subsection{Concept Drift} 

Concept drift specifically refers to a change in the underlying relationship between input features and the target variable \cite{gama2014survey}. Over time, the statistical properties of a target domain undergo arbitrary changes, introducing unpredictability into the learning process. Initially proposed by Schlimmer and Granger \cite{schlimmer1986incremental}, concept drift highlights the potential transformation of noise data into valuable non-noise information at different times. Concept drift manifests as past data patterns becoming irrelevant to new data, resulting in degraded predictions and decision outcomes, thereby decreasing effectiveness in model prediction efficacy within data-driven systems such as information systems and decision support systems that utilize these forms of inference \cite{liu2017regional}.

\begin{definition}
  (Concept Drift). Given a time period $[0,t]$, a set of samples, denoted as $S_{0,t} = \{d_0,\cdots,d_t\}$, where $d_i = (X_i, y_i)$ is one observation (or
a data instance), $X_i$ is the feature vector, $y_i$ is the label, and $S_{0,t}$ follows a certain distribution $F_{0,t} (X,y)$. Concept drift
occurs at timestamp $t+1$, if $F_{0,t} \neq F_{t+1,\infty} (X,y)$, denoted as $\exists t : P_t(X,y) \neq P_{t+1}(X,y)$ \cite{lu2016concept}, \cite{gama2014survey}.
\end{definition}

\subsection{ML Model Attack Accuracy and the Privacy Drift}

The connection between changes in attack accuracy in ML and privacy drift is significant. These changes typically stem from data updates or model modifications, leading to potential increases in the accuracy of privacy breaches such as MIA. Consequently, over time, privacy protections may weaken. Higher attack accuracy serves as an indicator that the model divulges more information about its training data, highlighting a drift in privacy safeguards and elevating the risk of compromising sensitive information. Specifically, as MIA attack accuracy rises, it signifies that the model progressively reveals more details about its training data, thus emphasizing the presence of privacy drift.

Consequently, a higher MIA attack accuracy reflects a weakening of privacy safeguards, making it easier for attackers to infer the presence of specific data points in the training set. Monitoring MIA attack accuracy helps detect and address privacy drift, ensuring that models maintain strong privacy protections.

The major factors affecting \textit{privacy drift} can be summarized as follows: 

\begin{itemize}
    \item \textit{Data Drift:} As new data is collected and added to the system, the nature of the data might change in ways that affect privacy. For example, combining datasets can increase the risk of re-identifying individuals even if the individual datasets were anonymized.
    
    \item \textit{Model Evolution:} Updates or changes to the machine learning model can inadvertently affect privacy. For instance, a model might learn to infer sensitive attributes that were not explicitly part of the training data.
    
    \item \textit{Drift in Inference Attack Accuracy:} Over time, adversaries might develop new techniques to extract sensitive information from models, known as model inversion or membership inference attacks. As these techniques evolve, the risk to privacy increases.
    
    \item \textit{Transfer Learning:} The way models are used can change over time, potentially leading to privacy issues. For instance, applying a model trained on one population to a different population might reveal unexpected patterns that compromise privacy.
    
\end{itemize}
\vspace{0.2cm}
\textbf{Synthesis:} The concepts of data drift \cite{ackerman2021automatically} and concept drift \cite{schlimmer1986incremental} are critical in understanding the challenges faced by ML models in dynamic environments \cite{gama2014survey}. Data drift occurs when the distribution of input data changes over time \cite{moreno2012unifying}, while concept drift refers to changes in the relationship between input features and target variables \cite{schlimmer1986incremental}. Both phenomena can lead to significant degradation in model performance and require adaptive strategies to maintain accuracy. Similarly, the notion of privacy drift, introduced in this paper, underscores the evolving risks to data privacy as models are incrementally trained. By examining these interconnected concepts, we can develop more resilient and privacy-preserving ML systems that adapt to changing data landscapes while safeguarding user information.

\begin{figure}[!htb]
    \centering
    \includegraphics[width=\linewidth]{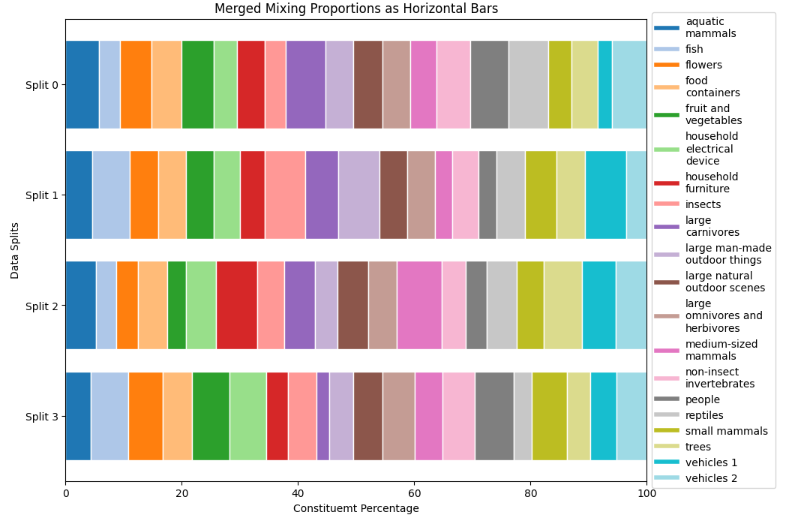}
    \caption{Four-split Non-IID  partitioning of CIFAR-20 Dataset.}
    \label{fig:cifar20abcd}
    \vspace{-10pt}
\end{figure}

\section{Threat Model} \label{threat}


This present the core threat model, providing an in-depth view of both the attacker's and the defender's perspectives \cite{rezaei2023accuracy}, \cite{shokri2017membership}.

\subsection{Defender's Perspective}


\textbf{Defender's  assumptions:}
We consider an FL environment where the defender leverages the training dataset by assigning a disjoint subset of training samples to each participating client. A sample is considered a member if it is included in the training data of at least one client. This decentralized approach confirms that each client operates on its own distinct portion of the dataset while collectively contributing to the model without direct data sharing.


In the context of multi-class categorization, the defender offers application programming interface (API) access and returns prediction confidence values. The ability to train base models from scratch enables the investigation of privacy-enhancing techniques, adjustments to the training process, and exploration of alternative fusion methods beyond traditional confidence averaging used in deep ensembles.


\textbf{Defender's  objectives:}
Within the FL framework, the primary goal of the defender is to reduce the potency of MIA while harnessing the accuracy advantages of ensemble learning. Throughout the entire process of training and inference, one of the primary concerns is the reduction of computational expenses. The goal of the study is to improve privacy protection in the FL paradigm while minimizing the loss of prediction accuracy.


\subsection{Adversary’s Perspective}


\textbf{Adversary’s knowledge:} The model architecture and training objectives are universally shared among the FL participants, making it a realistic assumption that the attacker has access to this level of knowledge. However, under the current assumption, the adversary has only restricted information about the target model and about the target dataset's distribution. This transparency allows the adversary to understand the general framework used in the learning process. The adversary cannot get knowledge about the global training process (federated or centralized) or how the training data is distributed among the clients.


\textbf{Adversary’s goals:} The adversary seeks to infer details about the data from the initial training set. Following the training phase, they create an attack model to deduce private information by querying the target model. The attacker does not modify the model’s parameters and relies solely on the data obtained through these queries, without needing additional information. Their focus is entirely on leveraging the responses from the model to deduce sensitive details  \cite{shokri2017membership}.


\textbf{Adversary’s  Capability:} We assume the adversary is honest but curious, meaning they can send queries to the target model but cannot access or modify its weights and gradients of the trained model. Despite these limitations, the adversary can still utilize query access to the target model to potentially launch MIA. Their goal is to infer whether a specific record or data point was part of the training dataset.

\begin{figure}[!ht]
    \centering
    \includegraphics[width=1\linewidth]{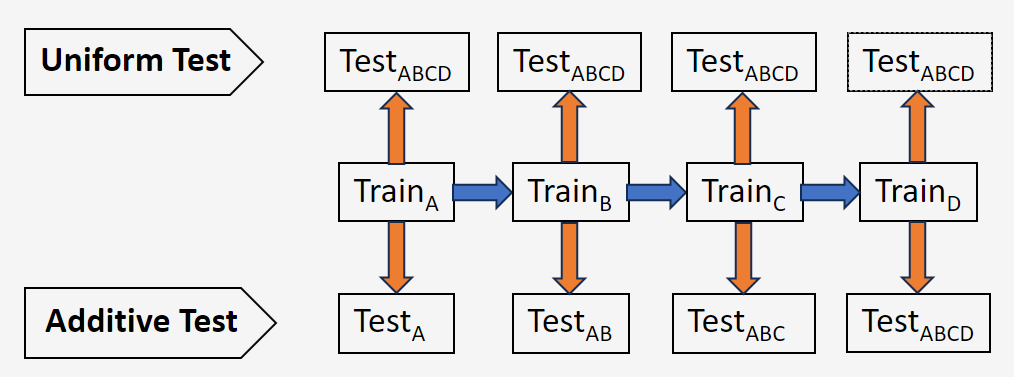}
    \caption{Design of training and test sets for incremental learning.}
    \label{fig:Experimental_Design}
\end{figure}

\section{Privacy Drift in FL: Dynamics and Trade-offs} \label{analysis}


In \cite{ahamed2024accuracy}, Ahamed \textit{et al.} established the inherent trade-off between model accuracy and privacy within an FL framework, particularly in the context of membership inference attacks (MIA). They further demonstrated that this trade-off is not correlated with the number of clients involved in the FL process. Building on these findings, we explore the concept of ``privacy drift" in this paper, investigating how incremental training and data distribution shifts impact privacy over time.

Our experiments reveal that privacy drift manifests as fluctuations in the susceptibility of models to privacy attacks during incremental updates. As the model undergoes continuous training with new data, the leakage of private information, as measured by the accuracy of MIA, varies. This indicates that privacy risks are dynamic and evolve alongside model performance.

We observe that improvements in model accuracy, achieved through various confidence-metric-based fusion strategies and other techniques, often correlate with increased privacy risks. This reinforces the accuracy-privacy trade-off identified in our previous work. However, the relationship between the number of clients and privacy drift remains complex and non-monotonic, suggesting that other factors, such as data distribution and model updates, play more significant roles in influencing privacy.

Through our detailed quantitative analysis, we identify key factors contributing to privacy drift, including data drift, model evolution, and changes in attack accuracy over time. These insights highlight the need for adaptive and robust privacy-preserving techniques in FL systems to manage privacy risks effectively as models evolve. Our findings, which are presented in Section \ref{results}, lay the groundwork for developing strategies to mitigate privacy drift, ensuring that FL systems can maintain a balance between accuracy and privacy in dynamic environments.

\section{Result Analysis and Discussion} \label{results}

In this section, we first define our experimental design and setup. Then, we demonstrate privacy drift and its trade-off with model accuracy under two different incremental learning paradigms. Finally, we investigate privacy drift in FL environments with varying numbers of clients.

\begin{figure*}
    \centering
    \includegraphics[width=\linewidth]{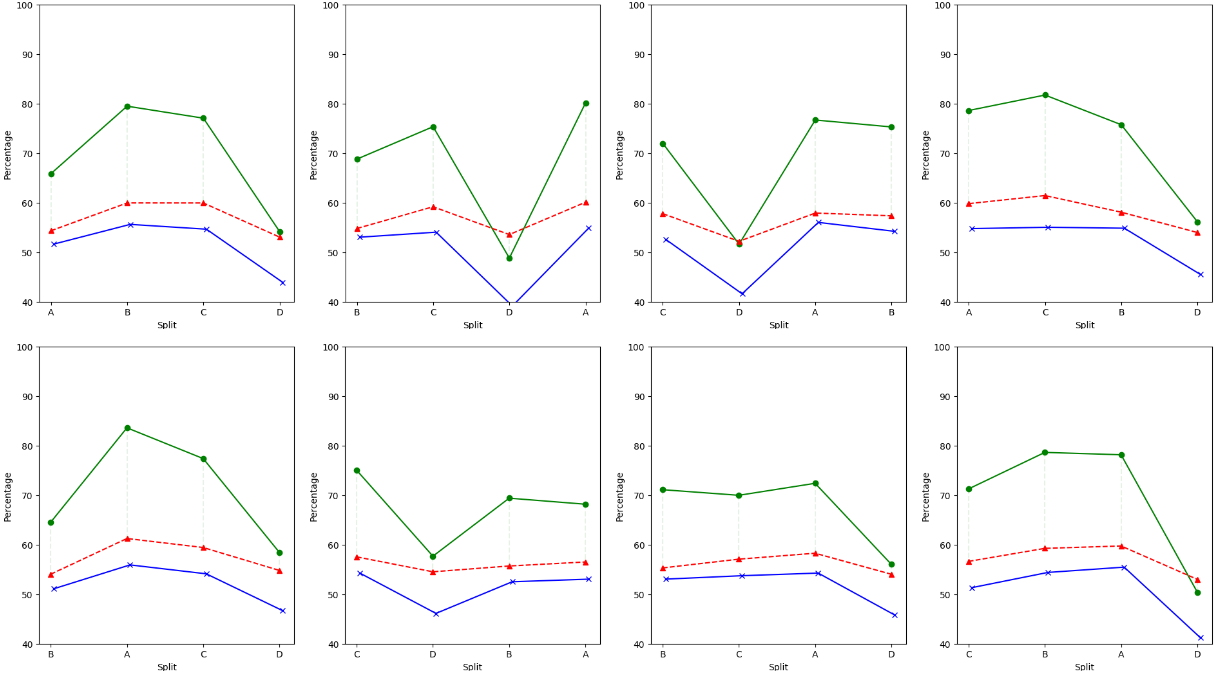}
    \caption{Training accuracy, test accuracy, and MIA AUC (area under the curve) for each permutation in the uniform test paradigm, illustrating the relationship between model performance and privacy leakage.}
    \label{Centralized_Uniform_All}
\end{figure*}


\subsection{Experimental Design}
For our experiments, we consider the CIFAR-100 dataset. As CIFAR-100 has 20 superclasses, and each superclass consists of 5 related classes, we effectively reduce CIFAR-100 to CIFAR-20. This reduction facilitates easier classification and faster training, supporting enhanced experimental flexibility. 
The original data is divided into four mutually exclusive partitions which are intermixed to build the CIFAR-20 dataset. The four partitions are referred as A, B, C, and D, respectively. These four splits follow Non-Independent and Identically Distribution (non-iid). \Cref{fig:cifar20abcd} demonstrates the distribution of constituent classes for the four splits of the CIFAR-20 dataset.

The motivation for this approach is to create a controlled environment where we can systematically analyze privacy drift. By splitting the data into four partitions (A, B, C, and D) through random sampling from the same underlying dataset, we simulate a real-world scenario where a model is continuously trained with new, non-iid data as it arrives. This setup allows us to observe how incremental learning and data distribution shifts impact privacy and model performance over time, reflecting the dynamic nature of incremental training environments.

\begin{figure}
    \centering
    \includegraphics[width=\linewidth]{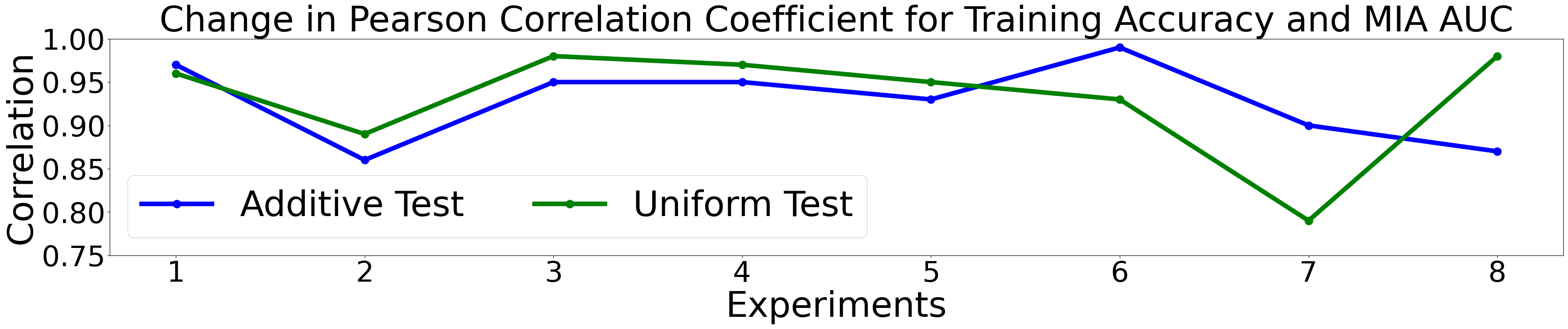}
    \caption{Pearson correlation between training accuracy and MIA AUC (area under the curve), illustrating the relationship across different data distributions and testing paradigms.}
    \label{Centralized_Correlation}
\end{figure}

The overall experimental design setup is depicted in \Cref{fig:Experimental_Design}. For incremental learning, we employ a sequential training approach. Initially, a model is trained on one partitioned dataset, designated as $X$. For each partitioned dataset $X \in S$, where $S = \{A, B, C, D\}$, we create a training set $Train_X$. The model is trained using $Train_X$, and the pre-trained model, and then this trained model is further trained on the next dataset's training set. This iterative process is repeated for each partitioned dataset, enabling the model to undergo incremental learning across various non-iid datasets.

The model's training performance is evaluated using two distinct testing methodologies: uniform test and additive test. In the uniform test, test datasets $Test_{X}$ from each partitioned dataset $X \in S$ contribute equally to form a single uniform test set, which is used throughout each of the training phases to ensure a stable benchmark for assessment. In contrast, the additive test employs an incremental testing strategy, with each phase adding a new set to the test set.

\Cref{fig:Experimental_Design} demonstrates that for training set sequence $Train_A$, $Train_B$, $Train_C$ and $Train_D$,  the test set for the uniform test is always  $Test_{ABCD}$. However, for the additive test paradigm, the test set becomes $Test_{A}$, $Test_{AB}$, $Test_{ABC}$ and $Test_{ABCD}$ respectively.
 
For training ML models over multiple clients in a FL setting,  we utilized the Nvidia Flare \cite{FLARE} and our code and experiments are publicly available \footnote{https://github.com/soumyaxyz/Privacy-Preserving-Federated-Learning}.

\begin{figure}[!htb]
    \centering
    \begin{subfigure}[b]{.49\linewidth}
        \includegraphics[width=\linewidth]{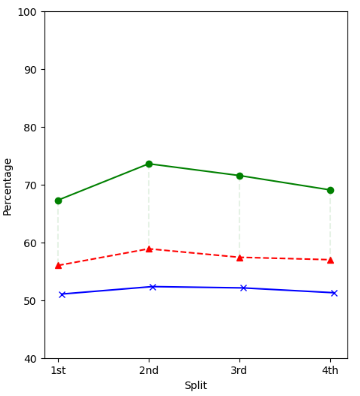}
        \caption{Centralized}
        \label{fig:Centralized_unif}
    \end{subfigure}
    \begin{subfigure}[b]{.49\linewidth}
        \includegraphics[width=\linewidth]{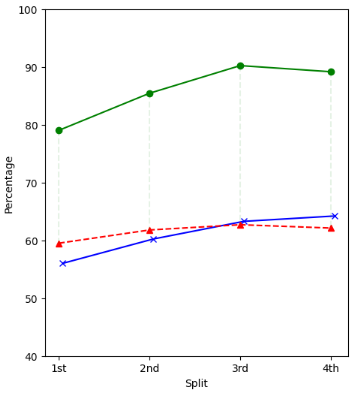}
        \caption{Two Client}
        \label{fig:Two_Client_unif}
    \end{subfigure}\\
    \begin{subfigure}[b]{.49\linewidth}
        \includegraphics[width=\linewidth]{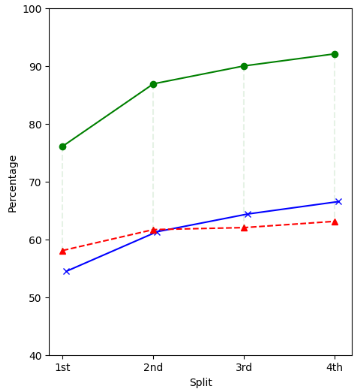}
        \caption{Five Client}
        \label{fig:Five_Client_unif}
    \end{subfigure}
    \begin{subfigure}[b]{.49\linewidth}
        \includegraphics[width=\linewidth]{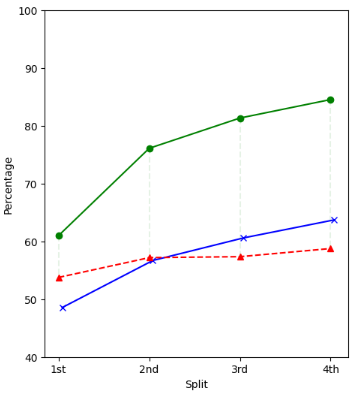}
        \caption{Ten Client}
        \label{fig:Ten_Client_unif}
    \end{subfigure}
    \caption{Privacy drift in CIFAR-20 under the uniform test paradigm.}
    
\end{figure}

\begin{figure}[!htb]
    \centering
    \includegraphics[width=\linewidth]{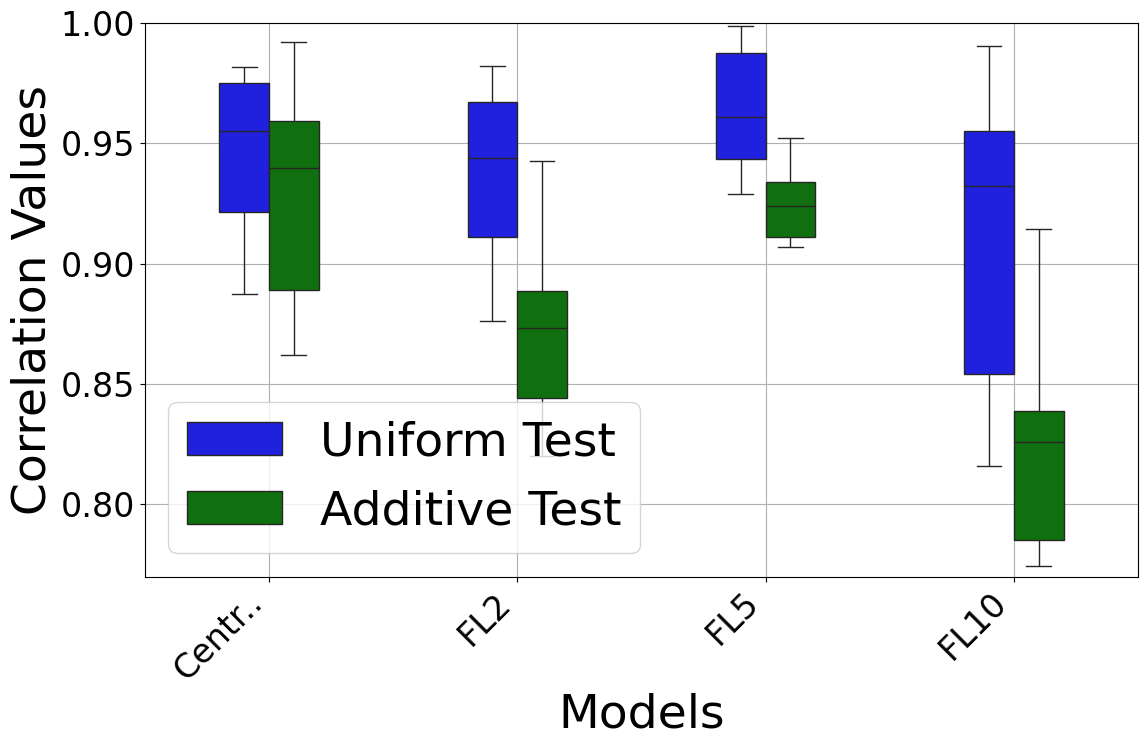}
    \label{ALL_cifar100_boxplot}
    \caption{Variation in the correlation for the individual experiments across different numbers of clients, for both uniform and additive test sets.}
    \label{fig:BoxPlot_Unif}
\end{figure}


\subsection{Privacy Drift with Incremental Training}

In this section, we investigate how a model's privacy evolves with incremental training using the EfficientNet architecture. We investigate different sequences (permutations) of the partitioned datasets, trying eight permutations, as a reasonable number of variations. We investigated the model both under uniform and additive test paradigms.  
\Cref{Centralized_Uniform_All} shows the training accuracy (green line), test accuracy (blue line), and MIA area under the curve (AUC) (red line) for each permutation in the uniform test paradigm. As expected, there is a strong correlation between training and test accuracy. Notably, we also observe a distinct correlation between training accuracy and MIA AUC, a proxy for privacy. Even in scenarios where training and test accuracy diverge, privacy remains correlated with training accuracy. This is because privacy is influenced only by the data the model has encountered.

\Cref{Centralized_Correlation} presents the Pearson correlation between training accuracy and MIA AUC. The Pearson correlation coefficient ranges from $-1$ to $1$: $1$ indicates perfect correlation, $0$ indicates no correlation, and $-1$ indicates perfect anti-correlation.
Regardless of the testing paradigm, the MIA AUC maintains a high correlation across different data distributions.

\subsection{Privacy Drift with respect to FL}

In this section, we investigate how privacy drift interplays with FL. Each split is equally and disjointly distributed between the FL clients and the model widths are aggregated through fed-average \cite{mcmahan2017communication}. We observe that that correlation between MIA AUC and training accuracy holds even when the training is distributed over different numbers of clients. However, there were a few interesting observations during this study. Firstly as an effect of our experimental design for high-client scenarios, the effective data points in a single split for a single client became too small for training in some cases. Especially, if we used stochastic gradient descent (SGD) \cite{SGD} as an optimizer in several of the cases the model failed to converge. For our experiments, we use the Adam Optimizer \cite{adam} which is more stable at low data paradigms. When we average the experimental results over the different data splits, we observed another effect of low training data, incremental training without sufficient data can be detrimental to the model's overall accuracy. However interestingly when we train the same model over the same data in a distributed Federated manner, the regularization effect of FL actually allows the model to learn with greater accuracy.

\begin{figure}[!htb]
    \centering
    \begin{subfigure}[b]{.49\linewidth}
        \includegraphics[width=\linewidth]{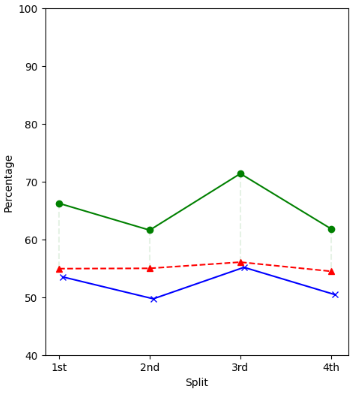}
        \caption{Centralized}
        \label{fig:Centralized_additive}
    \end{subfigure}
    \begin{subfigure}[b]{.49\linewidth}
        \includegraphics[width=\linewidth]{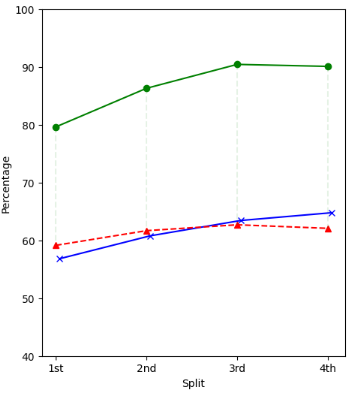}
        \caption{Two Client}
        \label{fig:Two_additive}
    \end{subfigure}\\
    \begin{subfigure}[b]{.49\linewidth}
        \includegraphics[width=\linewidth]{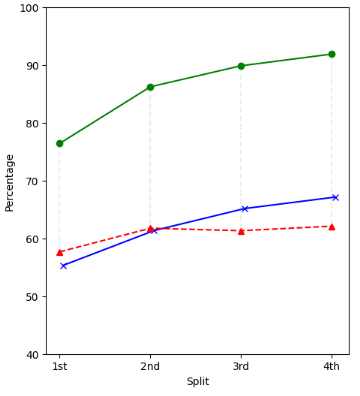}
        \caption{Five Client}
        \label{fig:Five_additive}
    \end{subfigure}
    \begin{subfigure}[b]{.49\linewidth}
        \includegraphics[width=\linewidth]{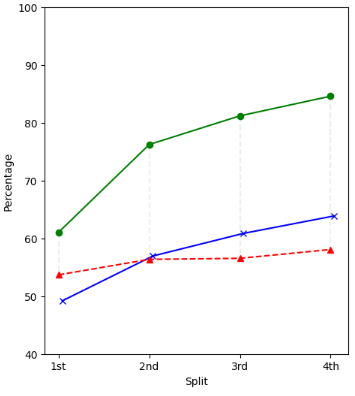}
        \caption{Ten Client}
        \label{fig:Ten_Client_additive}
    \end{subfigure}
    \caption{Privacy drift in CIFAR-20 under the additive test paradigm.}
    
\end{figure}

\Cref{fig:Centralized_unif} -- \Cref{fig:Ten_Client_unif} demonstrate the result for the average of the same experiment for centralized, two, five, and ten federated clients respectively. These results were generated by averaging the individual results for the uniform test paradigm across the eight differently permutated data sets. \Cref{fig:Centralized_additive} -- \Cref{fig:Ten_Client_additive} demonstrate the similar result for the additive test paradigm.

\Cref{fig:BoxPlot_Unif} 
 demonstrates the variation in the correlation for the individual experiments across the different numbers of clients, for the two testing paradigms. We observe that the phenomenon of privacy drift is independent of the number of clients in FL. And regardless of the experimental setting, there is a strong correlation between privacy and accuracy and it's interplay over incremental training.

\section{Conclusion and future Scope}\label{conclusion}
In this paper, we introduced the concept of ``privacy drift" to highlight the dynamic relationship between model accuracy and privacy leakage in FL systems. Our empirical analysis on CIFAR-100 datasets demonstrated how incremental training and data distribution shifts influence the susceptibility of models to privacy attacks, particularly membership inference attacks. We identified key factors contributing to privacy drift and provided a quantitative analysis of their impact.

Moving forward, future research can explore advanced privacy-preserving techniques to mitigate privacy drift in FL. This includes the integration of differential privacy, secure aggregation, and adversarial training methods. Additionally, extending the study to more diverse datasets and real-world applications will provide deeper insights into the practical implications of privacy drift. 

\section*{Acknowledgement}
This work is partially funded by the Deloitte AI Center of Excellence, and also supported in part by the Coastal Virginia Center for Cyber Innovation (CoVA CCI).

\bibliographystyle{IEEEtran}
\bibliography{reference.bib}
\end{document}